# Deep Learning Methods for Credit Card Fraud Detection

Thanh Thi Nguyen[1], Hammad Tahir[1], Mohamed Abdelrazek[1] and Ali Babar[2]

[1]School of Information Technology, Deakin University, Victoria, Australia
[2]School of Computer Science, The University of Adelaide, South Australia, Australia

E-mails: thanh.nguyen@deakin.edu.au, mohamed.abdelrazek@deakin.edu.au, ali.babar@adelaide.edu.au
Tel: +61 3 52278281

*Abstract*— **Credit card frauds are at an ever-increasing rate and have become a major problem in the financial sector. Because of these frauds, card users are hesitant in making purchases and both the merchants and financial institutions bear heavy losses. Some major challenges in credit card frauds involve the availability of public data, high class imbalance in data, changing nature of frauds and the high number of false alarms. Machine learning techniques have been used to detect credit card frauds but no fraud detection systems have been able to offer great efficiency to date. Recent development of deep learning has been applied to solve complex problems in various areas. This paper presents a thorough study of deep learning methods for the credit card fraud detection problem and compare their performance with various machine learning algorithms on three different financial datasets. Experimental results show great performance of the proposed deep learning methods against traditional machine learning models and imply that the proposed approaches can be implemented effectively for real-world credit card fraud detection systems.**

## I. Introduction

We live in a world today where with the power of a single touch we can achieve massive results. We can book rides, talk to personal virtual assistants, get recommendations, navigate maps and order food to our homes. All of this is possible only because of higher computing capabilities and shared IT infrastructure. Because of this, a large volume of data is created and data generation is expected to reach 44 zettabytes (40 trillion gigabytes) in 2020 from 4.4 zettabytes in 2013 [1]. The rise of artificial intelligence (AI) and machine learning (ML) in recent years is a consequence of this upsurge of data. Today we rely on countless implementations of ML in our everyday life without even realizing it. One such implementation is credit card fraud detection systems using ML techniques that make our payment methods more robust.

Credit card fraud is an extension of theft and fraud carried out by obtaining credit card details or using counterfeit cards for illegal monetary transactions. Credit card fraud can be physical where fraudster presents a credit card physically to the merchant or virtual where the transaction is made over the internet. Fraudsters use various techniques for this matter such as site cloning where a duplicate copy a merchant's website is created to obtain credit card information or cloning methods where magnetic stripe reader obtains information on the credit card to make a fake copy of it [2].

ML is a branch of AI in which a computer (machine) is able to perform predictions based on the findings from the previous data trends. Since the astonishing success of Google DeepMind in 2015, AI and ML have expanded to new horizons. Some practical implementations using deep anomaly detection are computer network intrusion detection [3]; banking, insurance, mobile cellular network, and health care fraud detection; medical and malware anomaly detection and anomaly detection for video surveillance [4]. Location tracking, Android malware detection, home automation and predicting the occurrence of heart disease are some applications of ML in the Internet of Things domain [5]. In this paper, we will study in-depth the practical implementation of ML, especially deep learning methods in detecting credit card frauds in the financial sector.

## II. Background and Literature

### A. Credit Card Frauds

In recent years, we have become more dependent on mobile phones and web applications, which have caused an increase in the number of online payment transactions. The card frauds have resulted in millions of revenue loss to financial institutions globally and added stress to credit card users. Approximately, 20.48 billion cards (including credit, debit and all prepaid cards) are in circulation worldwide for the year 2017 [6].

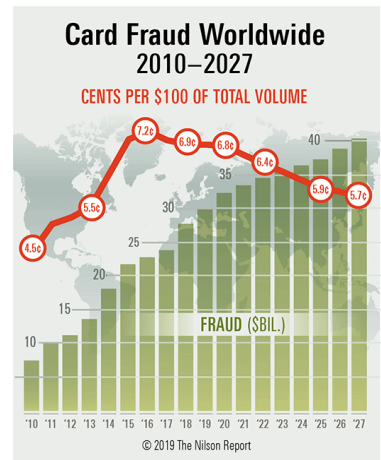

Fig. 1. Card Fraud Worldwide [7]

Fig. 1 provides the summary of gross card frauds and cents lost per 100 USD till 2027 globally [7]. It can be seen that credit card fraud reached to almost 30 billion USD in 2019 and is projected to increase each year while cents per 100 USD is expected to decrease. Australian Payments Network in their annual report shows an increase in credit card use from the previous years and a total of 574 million AUD lost to fraud in Australia. An increase to 10% among all credit card frauds was observed whereas the Card Not Present (CNP) fraud resulted in 84.9% of all frauds [8]. Some fraud limitation steps taken by Australian government include CNP Fraud Mitigation framework in which standards are defined for issuers and merchants, Australian Payment Council's partnership with the Joint Cyber-Security Centre to acquire actionable information in the said matter, regulation of EMV chips, avoiding refunds to alternative cards and implementing fraud detection systems by the financial institutions.

*B. Fraud Detection using Machine Learning*

ML can be implemented for credit card fraud detection where it becomes possible to classify an incoming transaction as legitimate or fraudulent based on the pattern of previous transactions [9]. ML, nature-inspired learning, and the combination of both these methods to form more robust hybrid models are used for card fraud detection [2]. Increased volume of card transaction data can be used to find outliers in the data by using methods like auto encoders, long short-term memory networks, and convolutional neural networks [4]. Decision tree classifier and ANN are also used for fraud detection and their performance is compared with rule-based models [10]. Decision tree classifiers are straightforward but their performance with complex data is quite low and ANNs perform better with large datasets but require heavy processing power. The rule-based methods are easy to implement but they are not good at classifying the new type of fraud.

The use of Predictive Analytic Technologies (PAT) to detect credit card frauds was advocated in [11]. PAT uses ML and statistical models to make future predictions. Five phases of the predictive analytics process are: outlining the business problem, acquiring and preparing data, analyzing data and formulating model, deploying predictive model and testing model performance. The common red flag schemes used by PAT to make predictions are uncommon purchase made by the cardholder, sudden identical purchases on the same credit card, purchases with overnight shipping, purchases with international shipment, multiple card shipments to a single address, multiple transactions on a card in short time, geolocation of transaction compared with cardholders registered location and usage of single IP address for multiple credit cards. Most of the vendor relies on ANNs for predictions but this method is restrained by a high number of false positives. Other challenges associated with PAT are model limitations such as implementation cost and complexity, limited training and learning competence and inability to adapt to fraud tactics, misclassification cost by emphasizing on large amount transactions and ignoring lower amounts commutatively, reluctance in sharing fraud data and compliance with law and regulation.

A method to detect fraudulent transactions with the least amount of false predictions was introduced in [12]. To do so, the methodology used is a combination of Hidden Markov Model (HMM), behaviour-based and genetic algorithms. HMM is used to generate transaction logs for a user and then a behaviour-based algorithm classifies each transaction into lower, medium and higher profiles by clustering the data and detects fraud by matching the transactions with the user's spending history. Then the genetic algorithm is used to calculate thresholds and at last fraudulent transaction is detected if the average value of all the three techniques is greater than 40%. Likewise, an experimental study for detecting credit card fraud using ML methods was provided in [13]. Out of eight classification algorithms tested C5.0 (decision tree algorithm), SVM, and ANN yielded promising results on a labelled dataset. To evaluate the performance combination of accuracy, recall and area under the precision-recall curve (AUC) are used. Accuracy alone is not used for evaluation because of the data imbalance in credit card frauds. The recall is the ratio of correctly identified fraud transactions over the actual number of fraud transactions. This ensures the robustness of the system. The shortlisted algorithms are then tested with imbalanced classification techniques: random over sampling, one-class classification, and cost-sensitive models.

A detailed mechanism using K-means clustering and the genetic algorithm to create new data samples for minority clusters to create a balanced dataset and improve classification performance was introduced in [14]. In that method, unsupervised learning is used to make clusters of similar data points. Then the genetic algorithm which is inspired by natural selection and genetics produced new samples for the minority classes. That method will help generate more balanced training sets for card fraud detection and classification error. In contrast, a study of ensemble learning to detect credit card frauds was reported in [15]. Ensemble learning is the approach used to combine several ML classifiers to increase prediction performance. ANNs and random forest correctly identify fraud and non-fraud cases. Misclassifying normal or fraudulent transactions are both associated with high financial cost. In efforts to reduce the number of misclassified instances, a combination of 3 feed-forward NNs with different hyperparameters and 2 random forest classifiers with 300 and 400 decision trees are used. The output is then calculated by taking the majority result for the 5 models.

A comparative study on simple NN, multilayer perception layer (MPL) and Convolutional Neural Network (CNN) was presented in [16]. The data used for this study is self-generated with 60000 transactions and 12 features. Features selected in generating this data are common attributes acquired from financial institution databases and usual predictors identified for predictive modeling. The dataset is highly imbalanced. The dataset is then balanced using under-sampling. The learning rate is set to 0.001 and the activation function used in this study is 'softmax'. The results showed that MPL performed best with the highest accuracy of 87.88% followed by CNN with an accuracy of 82.86%. Alternatively, results of a real-time deep learning model using auto-encoders for the credit card fraud detection was reported in [17]. The performance metrics selected for model evaluation are confusion matrix, precision, recall, and accuracy effectively. The non-linear auto regression predicted the most fraudulent transactions however it also misclassified most of the legitimate transactions. Logistic regression misclassified legitimate transactions the least but with low prediction accuracy for fraud cases. Under these circumstances, the deep NN Auto Encoder provides the most stable results with a relatively higher prediction rate and lower misclassification error. Likewise, CNNs were applied to detect credit card frauds in [18] because of its ability to reduce over-fitting and reveal hidden fraud patterns. That approach used feature engineering to generate aggregated features from the transaction data and introduce a novel feature called trading entropy. Synthetic fraud samples are created from real fraudulent data using cost-based sampling to balance the dataset. The sampled dataset is then transformed into a feature matrix based on different time windows. CNN similar to LeNet has 6 layers with input as a feature matrix. The evaluation parameter selected is the F1 score. The performance of different classifiers is increased when using the trading entropy feature and with comparison to NN, SVM, and RF, CNN produced a much better performance for all the different sample sets tested. Similarly, credit card fraud detection was investigated in [19] by looking at individual transactions and advocate using time in a sequence of same card transactions to capture the changing nature of fraud. In this regard, adding statistical features obtained from rea features can help improve classification performance. One example is transaction velocity that in a certain point of time will calculate number transactions carried out within a time frame. By adding time series components to the data, the authors compared the performance of SVM and LSTM models. The metrics set for the performance evaluation were AUC and mean squared error. The experimental results showed that LSTM performed far better than SVM in terms of evaluation metrics and also classification rate (transactions/sec).

A comparison between CNN, Stacked LSTM (SLSTM) and a hybrid model combining CNN and LSTM (CNN-LSTM) was presented in [20] for credit card fraud detection. CNN is powerful in learning from short term sequences in the data while LSTM is good in capturing long term sequences. The dataset used is from an Indonesian bank and the majority class of non-fraud values is under sampled in 4 different ratios to create 4 datasets for testing. This study represents features with respect to time and PCA is used for dimensionality reduction. The results revealed that increasing the ratio between non-fraud and fraud values increased the accuracy of the classifier. For training accuracy, SLSTM was on top, CNN-LSTM stood second and CNN was last in all 4 datasets. However, due to the imbalanced nature of the datasets, accuracy is not the only measure for performance validation and the AUC values reveal that CNN performed best after CNN-LSTM and then SLSTM, which highlights that the patterns for fraud transactions are subjugated by short-term relation over long-term.

III. PROPOSED DEEP LEARNING-BASED CLASSIFIERS

As ML methods rely solely on historical data, due to the sensitive nature of financial data protecting user's privacy, publicly available datasets are not very common, which limits the scope of study in this area. Because of this, every study in this area is limited to just one dataset if used any. In ML, performance of a model can differ widely for different datasets (business cases). How will performance vary for three datasets with a varying number of features and transactions is one research agenda in this study? Furthermore, credit card fraud detection methods face the problem of class imbalance as the number of fraud cases compared to a hundred thousand of normal transactions is very less. Aptly addressing class imbalance is a major challenge and how the behavior is changed by applying various sampling methods to deal with class imbalance is another aim of this study.

Traditional ML algorithms such as Support Vector Machines (SVM), Decision Tree (DT) and Logistic Regression (LR) have been extensively proposed for credit card fraud detection. These traditional algorithms are not very suitable for large datasets. The use of deep learning methods is still very limited and methods such as CNN and LSTM are encouraged for image classification and Natural Language Processing (NLP) respectively because of their ability to handle massive datasets. How theses deep learning methods perform for credit card fraud classification is the major focus of this study. In addition, data pre-processing is an important stage in the ML process. How the classification performance is affected in response to data pre-processing in credit card fraud detection is another question that needs to be answered.

*A. One-Dimensional CNN (1DCNN)*

CNN is a deep learning method heavily associated with spatial data such as image processing data. Similar to ANN, CNN has the same hidden layer structure in addition to special convolution layers with a different number of channels in each layer. The word convolution is linked with the idea of moving filters that capture the key information from the data. CNN is widely used in image processing as it automatically performs the feature reduction which makes it less prone to overfitting and thus training CNN does not require heavy data pre-processing. The role of using CNN for image processing is to minimize the processing by reducing the image without losing key features to make predictions [21]. The key terms in CNN are *feature maps, channels, pooling, stride*, and *padding*.

In comparison to the popular multi-layer perceptron (MLP) network, CNN are not fully connected in layer to layer connection and unlike MLP that has different weights associated with each node, CNN has constant weight parameter for each filter and these two features reduce the number of parameters in a CNN model. Also, the pooling method improves the feature detection process making it more robust to size and position changes of an element in an image.

CNN models are conventionally used for image and video processing that has two-dimensional data as input and therefore named as 2DCNN. The feature mapping process is used to learn the internal representation from the input data and the same procedure can be used for one-dimensional data as well where the location of features is not relevant. A very popular example of 1DCNN application is in Natural Language Processing which is a sequence classification problem. In 1DCNN, the kernel filter moves top to bottom in a sequence of a data sample instead of moving left to right and top to bottom in 2DCNN.

TABLE I. 2DCNN STRUCTURE

| Layer 1 | Input |
|---|---|
| Input Shape | (Row sample, 5, 6, 1) |
| Layer 2 | CONV2D |
| Number of channels | 64 |
| Kernel Size | 3x3 |
| Activation Function | ReLU |

| Layer 3 | CONV2D |
|---|---|
| Number of channels | 32 |
| Kernel Size | 3x3 |
| Activation Function | ReLU |
| **Layer 4** | **Flatten** |
| Number of Nodes | 64 |
| **Layer 5** | **Output** |
| Number of Nodes | 1 |
| Activation Function | Sigmoid |

TABLE II. 1DCNN STRUCTURE

| Layer 1 | Input |
|---|---|
| Input Shape | (Row sample ,1, Number of Features) |
| **Layer 2** | **CONV1D Layer** |
| Number of channels | 64 |
| Kernel Size | 1 |
| Activation Function | ReLU |
| **Layer 3** | **CONV1D Layer** |
| Number of channels | 64 |
| Kernel Size | 1 |
| Activation Function | ReLU |
| **Layer 4** | **Dropout** |
| Threshold | 0.5 |
| **Layer 5** | **MaxPooling1D** |
| Pool size | 1 |
| **Layer 6** | **Flatten** |
| Number of Nodes | 64 |
| **Layer 7** | **Dense** |
| Number of Nodes | 100 |
| Activation Function | ReLU |
| **Layer 8** | **Output** |
| Number of Nodes | 1 |
| Activation Function | Sigmoid |

This study uses both 2DCNN and 1DCNN to classify fraud and non-fraud cases. 2DCNN is used for the European Card Dataset in which the number of features is thirty. Each transaction sample is reshaped to a two-dimensional image and passed as input for the 2DCNN model. Tables I and II show the construction parameters of 2DCNN and 1DCNN respectively in this study.

### B. Long Short Term Memory Network

LSTM is a type of Recurrent Neural Network (RNN). A standard NN cannot keep track of the previous information and every time they have to perform the learning task from scratch. In very simple words, RNN is a neural network with memory. RNN tends to have short term memory because of the vanishing gradient problem. Backpropagation is the backbone of neural networks as it minimizes the loss by adjusting weights of the network that are found using gradients. In RNN, as the gradient moves back in the network it shrinks so the weight update is very small. The earlier layers in the network that are affected by this small update do not learn much and RNN loses the ability to remember early examples in long sequences making it a short-term memory network.

LSTMs come to the rescue for this short-term memory problem by having a cell state that is the memory of the network passing through each step and gates in each step that controls the flow of memory by keeping necessary and discarding irrelevant information. The gates used are the forget gate, input gate, and output gate. Forget gate decides what information to keep from the previous step, Input gate decides what information to add from the current step and the output gate decides the hidden state for the next step. The difference between the hidden state and cell state is that while the hidden state keeps the information of the model on previous inputs it still suffers from short term memory and the cell state overcomes that by remembering key information starting from the earliest examples in the sequence. To understand the complete flow of an LSTM cell, consider Fig. 2 below where each dotted box represents a single step [22].

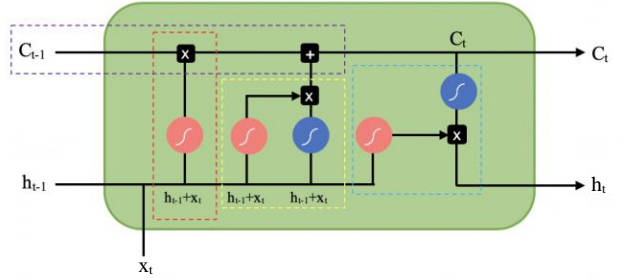

Fig. 2. An LSTM Cell

The first step shown in the red dotted box is the forget gate. Previous hidden state ($h_{t-1}$) and current input ($x_t$) are passed together to a ''sigmoid' activation function that provides the output between 0 (to forget) and 1 (to keep). The next step shown in the yellow dotted box is the input gate. In this gate, the previous hidden state and current input are passed through 'sigmoid' and 'tanh' activation function and the output of both is then multiplied. The 'tanh' function regulates the model while 'sigmoid' tell which information to keep from the current regulation. The next step shown in the purple dotted box calculates the cell state. Here the previous cell state ($C_{t-1}$) gets pointwise multiplied with the output of the forget gate and then the product is pointwise added with the output of the input gate to get the new cell state ($C_t$). The last step shown in the blue dotted box is the output gate that calculates the new hidden state ($h_t$). The new cell state is passed through 'tanh' activation function and is multiplied with the output of 'sigmoid' activation function which has an input of previous hidden state and current input. Table III below shows the hyperparameters of LSTM network used in this study.

TABLE III. LSTM STRUCTURE

| Layer 1 | Input |
|---|---|
| Input Shape | (1, Number of Features) |
| **Layer 2** | **Dense** |
| Number of LSTM Blocks | 50 |
| Activation Function | ReLU |
| **Layer 3** | **Output** |
| Number of Nodes | 1 |
| Activation Function | Sigmoid |

### IV. EXPERIMENTAL DATASETS

One main focus of this study is to determine the performance of classifiers for credit card fraud detection on datasets having a varied number of samples and features. For this purpose, three different datasets, i.e. European Card Data (ECD), Small Card Data (SCD) and Tall Card Data (TCD) are used. Like all credit card fraud datasets where fraud instances are very less compared to normal transactions, these datasets are highly imbalanced. All three datasets in this study are labelled datasets with class value '0' representing no fraud and '1' indicating fraud. Further details and class imbalance percentage is given below.

## A. European Card Data

This dataset courtesy of Machine Learning Group of Université Libre de Bruxelles is retrieved from Kaggle[1]. It contains two days of transaction data of European cardholders in September 2013. This dataset contains 284,807 samples and 31 features. Out of the given samples, only 492 are fraud cases and account for 0.172% of the dataset. Due to the privacy of customer information and the sensitivity of transactional details, all except 'Time' and 'Amount' features in the dataset are PCA transformed. The 'Time' feature represents the time in seconds passed starting from the first sample in the dataset and the 'Amount' feature shows the total amount of the transaction. This dataset is referred to as 'ECD' in this study.

## B. Small Card Data

This dataset, also retrieved from Kaggle[2], is a small dataset containing 3075 samples and 12 features. Half of the features are categorical while another half is numerical. Out of 3075 samples, 448 are fraud cases contributing to 14.6% of all cases. The features used in this dataset are Merchant ID, Transaction date, Average transaction amount per day, the Transaction amount is declined, Number of declines per day, is a foreign transaction, is a high-risk country, Daily chargeback average amount, six-month chargeback average amount, six-month chargeback frequency and is fraudulent. Due to a smaller number of rows and columns, this data set is named as Small Card Data (SCD).

## C. Tall Card Data

This dataset is obtained from an online database[3] courtesy of [13] containing 10 million samples (rows) and 9 features (columns). Having a high number of samples and a low number of features this dataset is named Tall Card Data (TCD) in this study. Only 5.96% of the dataset contains fraud cases. The features comprising the dataset are customer ID, gender, state, number of cards a customer has, balance on the card, number of transactions to date, number of international transactions to date, customer's credit line and fraud risk indicating fraud or non-fraud. Due to limited computing power and higher training times associated with the classifiers we take a small proportion of this data for the study. Out of 10 million samples, we only process 0.5 million samples which are almost double the samples in ECD. The class imbalance is kept the same and out of 500,000 total samples, there are 28,000 fraud cases.

## V. EVALUATION METRICS

Accuracy is not a suitable metric for model evaluation due to the high class imbalance in the datasets. Selecting the metric for evaluation depends on the nature of the solution and for credit card fraud detection systems capturing all fraud cases and reducing false alarm (legitimate transaction identified as fraud) is the goal. In this study, we use confusion matrix and denote non-fraud (legitimate) instances as Negatives and fraud instances as Positives. True negatives are the non-fraud cases predicted correctly, True positives are the fraud cases predicted correctly, False positives are the non-fraud cases predicted as fraud and False-negatives are the fraud cases predicted as non-fraud. To further understand the evaluation metrics, consider the equations for Accuracy, Precision, Recall, and the F1 Score given below.

$$Accuracy = \frac{TN + TP}{TN + TP + FN + FP} \quad (1)$$

$$Precision = \frac{TP}{TP + FP} \quad (2)$$

$$Recall = \frac{TP}{TP + FN} \quad (3)$$

$$F1 = 2 * \frac{Precision * Recall}{Precision + Recall} \quad (4)$$

It can be seen that precision is associated with positive predicted values. Decreasing the number of False positives will increase the precision so for circumstances where the cost of having False positives is high, precision is a suitable metric. Eq. (3) tells that recall is associated with actual positives. Decreasing the number of False negatives will increase the recall and problems with the high cost of False-negative tend to achieve high recall. For credit card fraud detection, a balance is required between False positives and False negatives. Predicting all the samples as fraud will have high recall but low accuracy, precision, and F1 score while predicting all samples as non-fraud will result in high accuracy but zero recall and undefined precision and F1 score. In this paper, we use all above four metrics for the comparisons.

## VI. ADDRESSING THE CLASS IMBALANCE PROBLEM

Class imbalance occurs when instances in a labelled dataset are not equally divided and the data is separated into majority and minority class/classes. Credit card transactional data is highly imbalanced because out of millions of transactions each day fraudulent transactions are very less. However, this does not mean that the impact of these fewer fraud cases compared to legitimate cases can be ignored. These fraud transactions heavily affect everyone in the card business, from user to merchant to issuer. Having an imbalanced dataset for problems where the class of interest is the minority class can lead to poor performance by having bias for majority class and misclassifying minority class by treating it as noise [23]. Data sampling, cost-sensitive learning, one-class learning, and ensemble learning are few methods to improve performance for imbalanced datasets. In this paper, under-sampling and over-sampling methods are further explored and performance on sample data for each dataset is evaluated for all the classifiers. The sampling methods used are as follow.

## A. Random Under Sampling

Random under sampling (RUS) is the process in which the instances from majority class are reduced by randomly selecting from the data. This is an under-sampling method where the class imbalance is reduced by reducing majority class. To perform random under sampling in python RandomUnderSampler class form imbalanced-learn library is used. The sampling ratio is selected by the 'sampling_strategy' parameter which is the ratio of required majority class over minority class.

---

[1] https://www.kaggle.com/mlg-ulb/creditcardfraud
[2] https://www.kaggle.com/shubhamjoshi2130of/abstract-data-set-for-credit-card-fraud-detection#creditcardcsvpresent.csv
[3] http://packages.revolutionanalytics.com/datasets/

### B. Near Miss Sampling

Randomly selecting instances in RUS can remove key information from the dataset and for this matter Near Miss (NM) sampling uses distance to select to sampling instances. Near miss has variants version 1, 2 and 3, and after evaluating performance of all three variants (results provided in section below), version 1 was selected in this study that selects the majority sampling instance that has the smallest average distance to the closest three instances of the minority class. In python the NearMiss class of imbalanced-learn library is used to perform this under sampling.

### C. Synthetic Minority Over Sampling Technique (SMOTE)

SMOTE is an over-sampling method that increases the number of instances in minority class by generating new synthetic samples. These new synthetic sample are generated by identifying nearest neighbors of the minority class sample and then generating a sample anywhere between the line of nearest neighbors. In this study SMOTE class of imbalanced-learn library is used to perform the over-sampling and the sampling ratio is the number of minority class instances after resampling over number of majority class instances.

## VII. EXPERIMENTS AND DISCUSIONS

### A. Data Pre-Processing

The first step in the experimentation is data pre-processing. In this step, all three data sets are explored in detail by enquiring the dataset manually and applying statistical operations. The purpose of data pre-processing is to provide a refined input to the classifiers to achieve best possible output. Missing values, categorical features, variable scale and high dimensionality can all affect the performance of the classifier. The two pre-processing methods involved in this study are data exploration, data scaling and test-train split.

Table IV down below provides key information on the data exploration process in this study. For ECD all the features were numeric, no missing value was found, and no feature was dropped to clean the data. All the categorical features in SCD were changed to numeric and 'Transaction Date' feature was dropped from the dataset as it was all of missing values. Both ECD and SCD datasets were used as whole while with TCD a smaller fraction of the actual dataset was used. For TCD all the values were numeric and 'custID' feature was dropped as it had all the unique values and added no information to the dataset.

Next in data exploration, correlation is distinguished between the features for each data set. Correlation is a statistical method and helps to establish the dependency of variables. It is a number between -1 to 1 where 0 means no relation at all, negative correlation means inversely proportional and positive correlation means directly proportional. Finding correlation can help eliminate the features that have the similar behaviour in the data reducing the dimesnions of the data. Lower dimensional data help improve training times and classification performance. Fig. 3 shows correlation between features in the SCD dataset. It can be observed that there is no negative correlation between the features and there is no uniform correlation throughout the SCD dataset. Higher average amount of transactions per day are associated with higher transaction amounts. Daily chargeback amount, six month chargeback amount and six month chargeback frequency are also highly associated with each other. Finally the high risk country feature is most significant in determining the fraud class. As this dataset is already very small no dimensionality reduction is performed on the data.

TABLE IV. DATA EXPLORATION

| ECD | |
|---|---|
| Number of Rows | 284807 |
| Number of Columns | 31 |
| Feature Type | Numeric |
| Missing Values | None |
| Dropped Features | None |
| Categorical to Numeric | None |
| Smaller Sample Used | No |
| SCD | |
| Number of Rows | 3075 |
| Number of Columns | 12 |
| Feature Type | Numeric + Categorical |
| Missing Values | 3075 |
| Dropped Features | 'Transaction date' |
| Categorical to Numeric | 'Merchant_id', 'Is declined', 'isForeignTransaction', 'isHighRiskCountry', 'isFradulent' |
| Smaller Sample Used | No |
| TCD | |
| Number of Rows | 10000000 |
| Number of Columns | 9 |
| Feature Type | Numeric |
| Missing Values | None |
| Dropped Features | 'custID' |
| Categorical to Numeric | None |
| Smaller Sample Used | Yes |

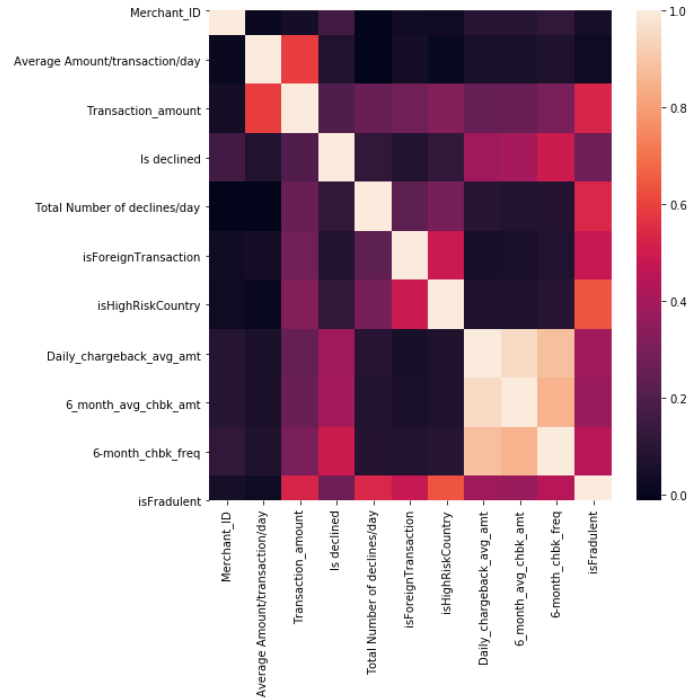

Fig. 3. Correlation Heatmap of the SCD dataset

## B. Data Scaling and Standardization

Normally the features in datasets are in different scale. Like the features 'Amount' and 'V1' in ECD dataset has mean 94826.6 and 0.000639 respectively. Deep learning algorithms does not perform very well if the input features are not on a fairly similar scale. Scaling and standardization methods bring the features together to almost same scale to make the input more comprehendible for the classifier. In this study, StandardScaler class from sklearn pre-processing library in python is used. The standard scaler transforms each feature in the dataset such that the mean is 0 and the standard deviation is 1. After applying the standard scaler on the dataset the mean for the same two features discussed above changed to 0.

## C. Test, Train and Validation Split

All three datasets are further divided into Test, Training and Validation data. Test Data is a small chunk of data obtained randomly from the dataset, which occupies 3.5% of each dataset. Training Data is the 80% of the remaining data used for training the models. Validation Data is the 20% of the remaining data used for validating the classifier. The classifiers used this validation data to avoid overfitting and improve model performance.

## D. Results and Discussions

Fig. 4 provides the comparison between different class imbalance ratios of the ECD dataset and the performance is evaluated using Test Dataset.

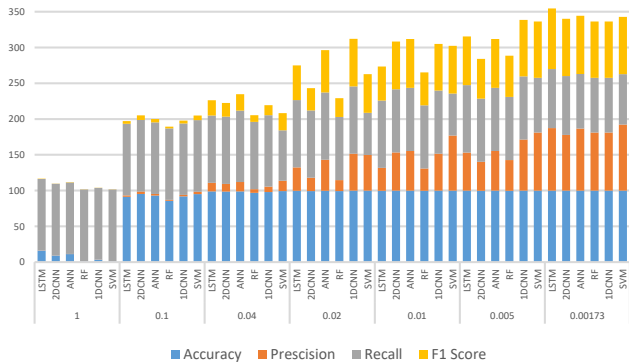

Fig. 4. Class Imbalance Comparison (Test Data ECD)

It can be seen that initially the Recall is maximum, and all other metrics are minimum and as the class imbalance is increased, recall begins to decrease and accuracy, precision and f1-score starts to increase. When there is no class imbalance at ratio 1, the model is predicting all the fraud cases correctly but misclassifying most of the non-fraud cases resulting in low precision. Low accuracy is achieved as the proportion of non-fraud cases is much larger than fraud cases in Test data. When the class imbalance starts to increase a significant rise in accuracy is observed meaning increase in correct predictions over total predictions. F1 score and Precision rise with increase in class imbalance. This is because with increase in class imbalance the model gets more training instances and is able to generalize well.

Fig. 5 shows performance comparison between the three datasets. The performance is calculated using both Validation and Test data and the performance measure selected here is F1-Score. It can be observed that in general validation performance is decreased as the dataset size is increased. SCD gives the best prediction performance on the examples that the system has already seen but drops down on the new unseen examples. TCD has the lowest validation and test performance amongst all dataset maybe due to the smaller number of features and little correlation between the features. The performance on ECD is stable with little variation between validation and test performance. The missing values of 2DCNN for SCD and TCD are because of the smaller number of features in the dataset as it was not possible to create feature matrix. Missing values of RF (i.e. random forest) and SVM in TCD are due to the higher training times associated with these classifiers.

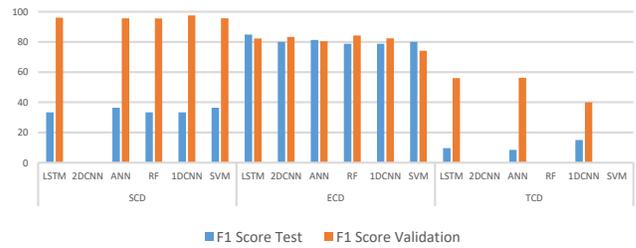

Fig. 5. F1 Score - SCD vs ECD vs TCD

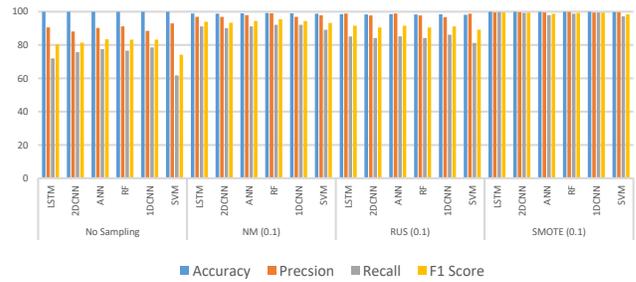

Fig. 6. Sampling Method Comparison (Validation Data ECD)

The comparison of sampling methods discussed earlier tested on Validation Data is presented in Fig. 6. It can be observed that overall performance is increased with sampling. SMOTE method provides the best results whereas Near Miss performs slightly better than Random Under Sampler. Finding the best sampling method leads to the next experiment.

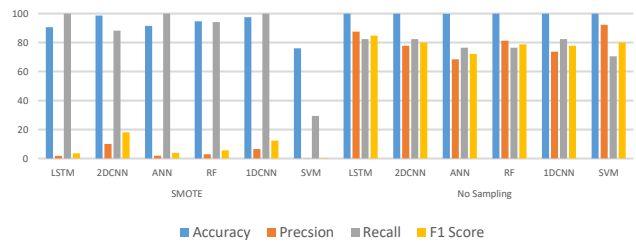

Fig. 7. SMOTE vs Normal Distribution (Test Data ECD)

Fig. 7 demonstrates the comparison of SMOTE with no sampling using the Test Data. As expected, the Recall is increased however Precision and F1-score are decreased drastically with SMOTE meaning the model is predicting fraud cases very accurately but misclassifying majority of the non-fraud cases. This may be because of the fact that SMOTE creates new synthetic fraud instances overlapping the non-fraud instances thus changing the decision boundary.

One of the main objectives of this study was to find the best performing algorithm. From Fig. 5, it can be observed that although deep learning methods performed side to side with the traditional algorithms, LSTM has a slightly better performance amongst all tested algorithms. Some other observations made during the experimentation process are as follows: 1) With the support of GPU computing, deep learning methods implemented based on the tensorflow library require less training time compared to traditional algorithms, i.e. SVM and RF, on large datasets; 2) Increasing the number of epochs increased the misclassification; 3) All the variants of Near Miss Algorithm provided the similar results.

## VIII. CONCLUSIONS AND FUTURE WORK

Credit card frauds are an increasing threat to the financial institutions. Fraudsters tend to come up with new fraud methods every now and then. A robust classifier is one that can cope with the changing nature of the frauds. Accurately predicting the fraud cases and reducing the number of false positives is the foremost priority of a fraud detection systems. The performance of machine learning methods varies for each business case. Type of input data is a dominant factor driving the machine learning model. For credit card detection, number of features, number of transactions and correlation between the features is an important factor in determining model performance. Deep learning methods such as CNN and LSTM are associated with image processing and NLP respectively. Using these methods for credit card fraud detection yielded better performance than traditional algorithms. While all the algorithms performed side to side, LSTM with 50 blocks was the one on top with F1-Score of 84.85%. In this study, sampling methods have been used to deal with the class imbalance problem. Using various sampling methods increased the performance on existing examples but decreased it significantly on the newly unseen data. The performance on unseen data was increased as the class imbalance was increased. Future work associated with this study is to explore hyperparameters used to construct deep learning methods to improve model performance.